\documentclass{ecai} 
\usepackage{latexsym}
\usepackage{amssymb}
\usepackage{amsmath}
\usepackage{amsthm}
\usepackage{booktabs}
\usepackage{enumitem}
\usepackage{graphicx}
\usepackage{color}

\usepackage{multirow}
\usepackage{array}
\usepackage{amsfonts}
\usepackage{marvosym}
\usepackage{algorithm}
\usepackage{algorithmic}
\usepackage{graphicx}
\usepackage[table]{xcolor}

\newcommand{\BibTeX}{B\kern-.05em{\sc i\kern-.025em b}\kern-.08em\TeX}

\begin{document}

%%%%%%%%%%%%%%%%%%%%%%%%%%%%%%%%%%%%%%%%%%%%%%%%%%%%%%%%%%%%%%%%%%%%%%%%

\begin{frontmatter}

%%% Use this command to specify your submission number.
%%% In doubleblind mode, it will be printed on the first page.

\paperid{700} 

%%% Use this command to specify the title of your paper.

\title{Generalized Face Anti-spoofing via Finer Domain Partition and Disentangling Liveness-irrelevant Factors}

%% Use this combinations of commands to specify all authors of your 
%% paper. Use \fnms{} and \snm{} to indicate everyone's first names 
%% and surname. This will help the publisher with indexing the 
%% proceedings. Please use a reasonable approximation in case your 
%% name does not neatly split into "first names" and "surname".
%% Specifying your ORCID digital identifier is optional. 
%% Use the \thanks{} command to indicate one or more corresponding 
%% authors and their email address(es). If so desired, you can specify
%% author contributions using the \footnote{} command.

\author[1]{\fnms{Jingyi}~\snm{Yang}}
\author[2]{\fnms{Zitong}~\snm{Yu}}
\author[3]{\fnms{Xiuming}~\snm{Ni}}
\author[3]{\fnms{Jia}~\snm{He}}
\author[1]{\fnms{Hui}~\snm{Li}\thanks{Corresponding Author. This work was supported by the National Science Foundation of China, under Grant No. 62171425 and Guangdong Basic and Applied Basic Research Foundation (Grant No. 2023A1515140037).}}

\address[1]{Dept. EEIS, University of Science and Technology of China \\ The CAS Key Laboratory of Wireless-Optical Communications}
\address[2]{School of Computing and Information Technology, Great Bay University}
\address[3]{Anhui Tsinglink Information Technology Co.,Ltd.}

%%% Use this environment to include an abstract of your paper.

\begin{abstract}
Face anti-spoofing techniques based on domain generalization have recently been studied widely. Adversarial learning and meta-learning techniques have been adopted to learn domain-invariant representations. However, prior approaches often consider the dataset gap as the primary factor behind domain shifts. This perspective is not fine-grained enough to reflect the intrinsic gap among the data accurately. In our work, we redefine domains based on identities rather than datasets, aiming to disentangle liveness and identity attributes. We emphasize ignoring the adverse effect of identity shift, focusing on learning identity-invariant liveness representations through orthogonalizing liveness and identity features. To cope with style shifts, we propose Style Cross module to expand the stylistic diversity and Channel-wise Style Attention module to weaken the sensitivity to style shifts, aiming to learn robust liveness representations. Furthermore, acknowledging the asymmetry between live and spoof samples, we introduce a novel contrastive loss, Asymmetric Augmented Instance Contrast. Extensive experiments on four public datasets demonstrate that our method achieves state-of-the-art performance under cross-dataset and limited source dataset scenarios. Additionally, our method has good scalability when expanding diversity of identities. The codes will be released soon.
\end{abstract}
\end{frontmatter}
% , which align the distribution of each domain based on domain labels
%%%%%%%%%%%%%%%%%%%%%%%%%%%%%%%%%%%%%%%%%%%%%%%%%%%%%%%%%%%%%%%%%%%%%%%%

\section{Introduction}
Face recognition (FR) technology has been widely used in various applications such as access control systems and mobile payments. Unfortunately, FR systems are vulnerable to various presentation attacks, including printing attacks, video replays, and 3D masks among others. To counter such risks, researchers have proposed face anti-spoofing (FAS) techniques, which have become a popular research topic in recent years. A series of face anti-spoofing methods have been developed, including those based on hand-crafted features \cite{peixoto2011face,de2013lbp,patel2016secure,boulkenafet2016face}, as well as deeply-learned features \cite{feng2016integration,yang2014learn,li2016original}, both of which have shown promising results in intra-dataset scenarios. However, they often exhibit poor generalization ability to unknown domains, largely due to assumptions about the stationary settings of liveness-irrelevant factors such as lighting, resolution of capture devices. They struggle to overcome dataset bias (domain shifts) resulting from real-world scenarios. Consequently, a significant challenge for FAS systems is the inability to effectively transfer the anti-spoofing models, learned within one or several domains, to an unseen domain.

% Consequently, FAS systems confront the challenge problem is that anti-spoofing models learned from source domains cannot be transferred to target domains consistently. 

\begin{figure}[t!]
\includegraphics[width=0.49\textwidth,height=0.22\textwidth]{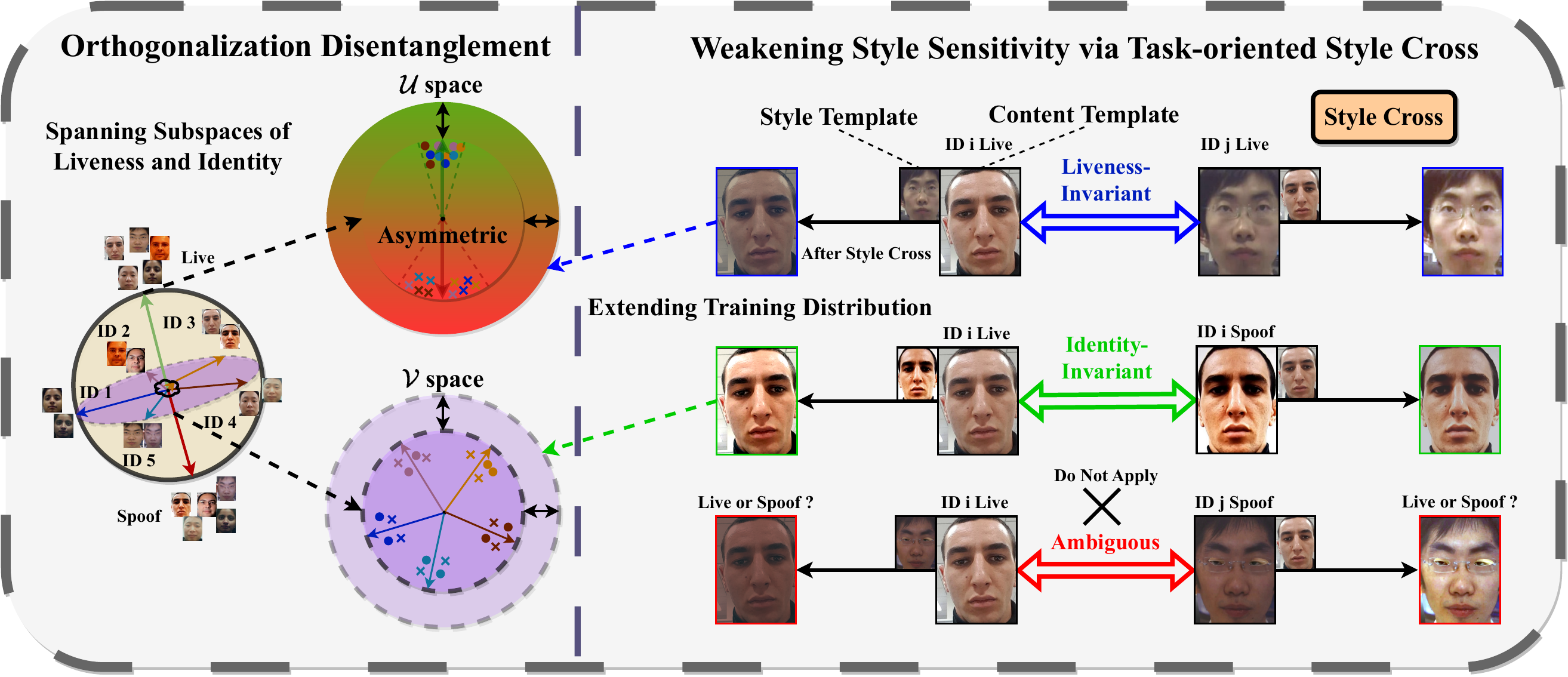}
\caption{\textbf{(Left)} Orthogonalization of liveness and identity attributes. The earth's axis represents the subspace $\mathcal{U}$ associated with the liveness component, where the green and red arrows indicate "live" and "spoof". The equatorial plane represents the subspace $\mathcal{V}$ belonging to the identity component, and colored arrows represent different identities. \textbf{(Right)} In $\mathcal{U}$ space, the liveness of the content template and the style template should be consistent. While in $\mathcal{V}$ space, the identity invariance is guaranteed.}
\label{fig: U V space}
\vspace{0.5cm}
\end{figure}

% A seemingly similar idea is also mentioned in \cite{tu2020learning}, however, a key distinction lies in their use of a weight-shared model for extracting identity and liveness features without an explicit disentangling process. They aim to solve both FAS and FR tasks simultaneously through a single network, potentially may result in both tasks making concessions to each other. Additionally, they employ domain adaptation (DA) techniques, assuming that the unlabeled target domain data is visible during training. In contrast, our work belongs to the DG family, where the target domain data remains unseen during training making it a more challenging scenario. 

To overcome the domain shifts, recent studies have leveraged domain generalization (DG) techniques to enhance the generalization ability of FAS system to unknown domains. The majority of DG-based FAS (DG-FAS) approaches utilize adversarial learning \cite{li2018domain,shao2019multi,jia2020single} or meta-learning \cite{shao2020regularized,liu2021adaptive} to learn domain-invariant representations. However, these approaches consider the dataset gap as the intrinsic divergence among the data and employ dataset partition as domain. The domain labels they used are coarse, and cannot comprehensively reflect the latent correlation from data. Because, even within the same dataset, inconsistencies, such as identities, illuminations, and resolutions may exist. Though D\(^2\)AM \cite{chen2021generalizable} attempts to assign pseudo-domain labels through clustering, it roughly aggregates the source data into a few clusters which does not solve the problem in essence. Moreover, another considerable drawback is that when the number of source datasets is limited, it is not conducive to learning domain-invariant representations. Under the limited source dataset scenarios, the performance of dataset partition is significantly inferior. In extreme situations, such methods are powerless when only one source dataset is available.

In this paper, we refine the factors that cause the domain shift into identity, style, and unseen spoof patterns, rather than vague dataset gaps. Additionally, we adopt a more finer domain partition according to identities instead of datasets. Our framework, named \textit{\textbf{D}isentangling \textbf{L}iveness-\textbf{i}rrelevant \textbf{F}actors} (\textbf{DLIF}). Concretely, we employ two networks to extract liveness and identity features separately. These features are then treated as dissimilar and expressed as orthogonal from a subspace perspective, instead utilizing generative adversarial network and pixel reconstruction approaches like \cite{wang2020cross,yue2022cyclically}, which exhibit heavy computational overheads. To enhance the efficiency and scalability of our framework, we propose two plug-and-play modules: Style Cross (SC) and Channel-wise Style Attention (CWSA). Specifically, SC is a feature-level style augmentation technique, and we explore in detail the effectiveness of executing it at different levels of the network. Meanwhile, in order to prevent label ambiguity caused by uncontrolled random SC in specific tasks, we implement Liveness-invariant Style Cross for FAS network and Identity-invariant Style Cross for FR network. CWSA, designed to adaptively generate style-insensitive features based on channel styles, is introduced specifically for FAS to further mitigate the impact of style shifts. Furthermore, we propose a Asymmetric Augmented Instance Contrastive loss which consider the asymmetry of live and spoof samples to learn robust liveness representation distribution. Simultaneously, our method exhibits excellent scalability. Building on the success of face recognition, we can leverage the knowledge gained from well-trained FR networks to provide an auxiliary supervision for FAS networks. Our main contributions are four-fold:
\begin{itemize}
\item We propose a novel perspective involving finer domain partition corresponding to identity. Here, liveness features and identity features are considered orthogonal and disentangled through orthogonality, as illustrated in Figure \ref{fig: U V space} (Left).
\item We propose a plug-and-play Style Cross module for style augmentation, as shown in Figure \ref{fig: U V space} (Right), along with a Channel-wise Style Attention module to learn style-insensitive features. 
\item We propose an Asymmetric Augmented Instance Contrast loss that asymmetrically treats live (homogeneity-aware) and spoof (heterogeneity-aware) instances which substantially improves the generalization capability.
\item Our framework is compatible with most existing face recognition models, enabling us to leverage well-trained face recognition models for disentangling rather than training from scratch. The scalability study demonstrates that after utilizing the well-trained FR models, there is a significant improvement in performance. Furthermore, if we increase the identity diversity of the training data, the performance will be further improved.
\end{itemize}

\begin{figure*}[t]
\includegraphics[width=0.97\textwidth,height=0.28\textwidth]{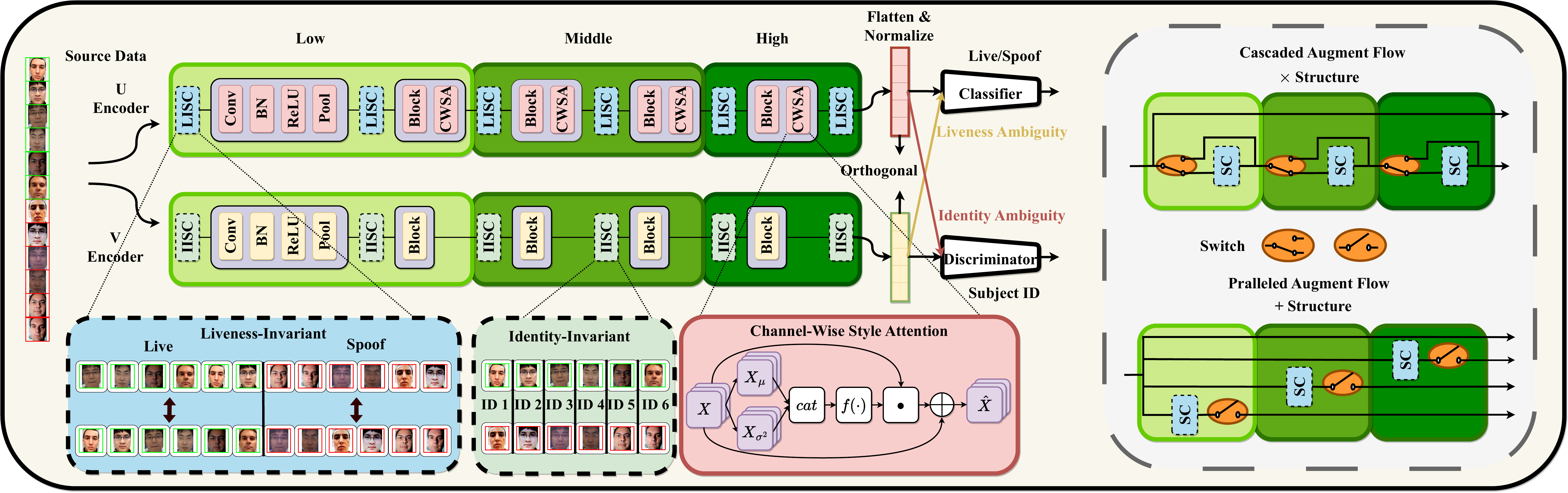}
    \caption{\textbf{(Left)} The architecture mainly consists of two encoders: encoder $U$ and $V$. $U$ extracts the liveness feature, and $V$ extracts the identity feature. The SC implements two types of mode in $U$ and $V$ which are liveness-invariant and identity-invariant, the dashed line indicates detachable, and we use colors from light to dark to represent the low, middle, and high levels of the encoder. The CWSA is utilized to weaken the sensitivity of the model for style variation. In addition, \textbf{(Right)} shows the style augmented flow of (\(\times\)) and (\(+\)) structures.}
    \label{fig: architecture}
\vspace{0.3cm}
\end{figure*}

\section{Related Work}
\paragraph{Face Anti-spoofing Methods.} Early handcrafted methods, such as SIFT \cite{patel2016secure}, LBP \cite{de2013lbp}, and HOG \cite{yang2013face} were utilized to address the FAS problem. Subsequently, with the emergence of CNN-based deep networks, binary classification-based approaches gain popularity \cite{yang2014learn,li2016original}, or leveraging auxiliary supervised signals that contain rich anti-spoofing information, such as pseudo-depth maps, reflection maps, and rPPG signals \cite{atoum2017face,liu2018learning}. Furthermore, novel custom operators \cite{yu2020searching,yu2020face} were introduced, suggesting their effectiveness for the FAS task. Recent efforts have delved into domain adaptation (DA) \cite{wang2020unsupervised,wang2021self}, and domain generalization (DG) \cite{jia2020single,liu2021adaptive,sun2023rethinking} to achieve generalization on unseen domains.

\paragraph{Generalizing Domain-specific Styles.} Previous work \cite{tang2021crossnorm} indicates that implementing feature-level style crossing can expand the training distribution and enhance the generalization ability against to domain shift. Inspired by AdaIN \cite{huang2017arbitrary}, SSAN \cite{wang2022domain} first proposed the shuffle style assembly (SSA) which randomly swap and mix source styles among source contents for the FAS problem. Additionally, Zhou et al. \cite{zhou2023instance} declare that the feature covariance stores domain-specific features, and instance whitening is effective in removing such domain-specific styles in image translation. They propose asymmetric instance adaptive whitening to align instances at a finer granularity. Those suggest that it is worthwhile to consider feature-level style augmentation and style sensitivity weakening methods that are more applicable to the FAS problem. 

\paragraph{Disentangling Liveness-Irrelevant Representation.} Disentangled representation learning (DRL) focuses on extracting features that can effectively capture correlations between different datasets, to solve the problem that features related to different tasks can be easily coupled with each other when there is no clear guidance. \cite{zhang2020face,yue2022cyclically} divide the representation of an image into content and liveness parts to solve FAS problems. Wang et al. \cite{wang2020cross} explicitly disentangles identity with liveness. Liu et al. \cite{liu2020disentangling} disentangles a spoof face into a live counterpart and spoof trace and aims to explicitly extract the spoof traces from faces. These methods entail pixel-level image reconstruction and adversarial training of GANs, which involves significant training cost and difficulty. Instead, we consider the liveness and identity to be dissimilar, a relationship that can be expressed as orthogonal in terms of cosine similarity. It is simpler and more efficient to execute.

\section{Proposed Method}
Our architecture is illustrated in Figure \ref{fig: architecture}. We employ encoder \(U\) for the FAS task and encoder \(V\) for the FR task, respectively. Both encoders employ ResNet18 as their backbone and are equipped with task-oriented style cross modules at different levels. First, the disentanglement of liveness and identity is introduced in Sec \ref{Orthogonalization Identity Disentanglement}. Next, we present the part on weakening style sensitivity in Sec \ref{Weakening Style Sensitivity}. Finally, the Asymmetry Augment Instance Contrast is detailed in Sec \ref{Asymmetric Augment Instance Contrast}.

\subsection{Problem Definition and Notations}
We have source (training) and target (testing) samples, denoted as \(S=\{(x_s^i,y_{fas}^i,y_{id}^i)\}_{n_s,n_{id}}^i,T=\{(x_t^i,y_{fas}^i)\}_{n_t}^i\), where \(x_s^i,x_t^i \in \mathbb{R}^{H\times W\times 3}\). \(y_{fas}^i\)is a one-hot label corresponding to liveness. \(y_{id}^i\) is a one-hot label corresponding to identity. \(n_{id}\) denotes the number of source subject IDs, and \(n_{s},n_{t}\) denotes the number of source, target samples. The goal of encoder \(U_\theta\) and classifier \(C_\theta\) is to categorize between live and spoof. The goal of encoder \(V_\theta\) and discriminator \(D_\theta\) is to assist \(U_\theta\) and \(C_\theta\) for disentangling. For simplicity, in cases where there is no ambiguity, we generally omit the subscript \(\theta\) of \(U_\theta\), \(V_\theta\), \(C_\theta\), and \(D_\theta\).

\subsection{Orthogonalization Identity Disentanglement}
\label{Orthogonalization Identity Disentanglement}
In our scheme of learning task-relevant representations, \(U\) generates the liveness feature $f_u=U(x)\in \mathbb{R}^N$ in space $\mathcal{U}$ for live/spoof classification, while \(V\) generates the identity feature $f_v=V(x)\in \mathbb{R}^N$ in space $\mathcal{V}$. Here $N$ represents the dimension of features. If we don't impose any constraints, space $\mathcal{U}$ and space $\mathcal{V}$ may be intertwined, and the $f_u$ and $f_v$ are not orthogonal. However, In this work, we assume that the liveness is irrelevant to the identity, they are treated as dissimilar and expressed as orthogonal through cosine similarity. Thus that we aim to orthogonalize all $f_u$ and $f_v$ by minimizing the square of cosine similarity between them. To construct $\mathcal{U} \perp \mathcal{V}$, a similarity matrix $M_{uv}$ for each batch is constructed as: 
\begin{equation}
M_{uv}=F_uF_v^T,
\end{equation}

Here, $F_u=\mathrm{Norm}(f^1_u, f^2_u, \cdots, f^B_u)^T\in \mathbb{R}^{B \times N}$  and $F_v=\mathrm{Norm}(f^1_v, f^2_v, \cdots, f^B_v)^T\in \mathbb{R}^{B \times N}$ where $B$ denotes the batch size, $\mathrm{Norm}$ denotes the normalization. The shape of $M_{uv}$ is $B\times B$ and $M_{uv}(i,j)$ represents the cosine similarity between $f^i_u$ and $f^j_v$. Accordingly, the orthogonal loss in each batch is defined as follows:
\begin{equation}
\label{eqn: loss ortho}
L_{ortho}=\frac{1}{B^2}\sum_{i=1}^B\sum_{j=1}^B\|M_{uv}(i,j)\|_2^2,
\end{equation}

Moreover, it is crucial for classifier \(C\) to exhibit liveness ambiguity (li-amb) towards \(f_v\), while discriminator \(D\) should demonstrate identity ambiguity (id-amb) towards \(f_u\). Consequently, we introduce two losses, \(L_{liamb}\) and \(L_{idamb}\), to account for these characteristics:

\begin{equation}
\label{eqn: loss idamb}
L_{{idamb}}=\frac{1}{B}\sum_i^B\left\|D(\mathrm{Norm}(U(x_s^i)))-\frac{1}{n_{id}}\right\|_2^2,
\end{equation}
\begin{equation}
\label{eqn: loss liamb}
L_{{liamb}}=\frac{1}{B}\sum_i^B\left\|C(\mathrm{Norm}(V(x_s^i)))-\frac{1}{2}\right\|_2^2,
\end{equation}

The above two loss functions ensure that the normal vector of hyperplane used for determining liveness property in space $\mathcal{U}$ is orthogonal to space $\mathcal{V}$. Likewise, the normal vectors of multiple classification boundaries in space $\mathcal{V}$, are orthogonal to space $\mathcal{U}$, as depicted in Figure \ref{fig: U V space}. For some succinct and intuitive theoretical interprets, please refer to the \textit{Supplementary Materials}.

\subsection{Weakening Style Sensitivity}
\label{Weakening Style Sensitivity}
Aiming to expand the diversity of style, we propose the task-oriented Style Cross (SC). In the context of the FAS task, the application of SC between live and spoof samples may introduce uncertainty regarding liveness. To address this concern, we limit the implementation of SC between samples with identical liveness labels, termed as Liveness-invariant Style Cross (LISC) in Eqn \ref{eqn: LISC 1}, \ref{eqn: LISC 2}. Similarly, for FR task, to alleviate potential ambiguities in identity, we adopt Identity-invariant Style Cross (IISC) between samples that share the same identity, in Eqn \ref{eqn: IISC 1}. Moreover, unlike shuffle style assembly \cite{wang2022domain} which involves parameter layers, we chose a lightweight design, directly exchanging styles channel by channel between samples. Style Cross is enabled during training and disabled during testing.\par

\begin{equation}
\label{eqn: LISC 1}
\mathrm{LISC}(f_{u(l)}^a,f_{u(l)}^b,\text{\small Livness=Live})=\sigma_{u(l)}^b \frac{f_{u(l)}^a - \mu_{u(l)}^a}{\sigma_{u(l)}^a} + \mu_{u(l)}^b
\end{equation}
\begin{equation}
\label{eqn: LISC 2}
\mathrm{LISC}(f_{u(s)}^a,f_{u(s)}^b,\text{\small Liveness=Spoof})=\sigma_{u(s)}^b \frac{f_{u(s)}^a - \mu_{u(s)}^a}{\sigma_{u(s)}^a} + \mu_{u(s)}^b
\end{equation}
\begin{equation}
\label{eqn: IISC 1}
\mathrm{IISC}(f_{v(i)}^a,f_{v(i)}^b,\text{\small ID}=i)=\sigma_{v(i)}^b \frac{f_{v(i)}^a - \mu_{v(i)}^a}{\sigma_{v(i)}^a} + \mu_{v(i)}^b
\end{equation}
where \(f_{u(l)}^a\) and \(f_{u(l)}^b\) represent two different live samples, however, \(f_{u(s)}^a\) and \(f_{u(s)}^b\) represent two different spoof samples. \(f_{v(i)}^a\) and \(f_{v(i)}^b\) represent two samples with the same identity. \(\mu\) and \(\sigma\) represent channel-wise mean and standard deviation respectively.

Additionally, we conduct an extensive investigation to determine the optimal level of models for performing SC, as well as the most effective augmentation flow. The right of Figure \ref{fig: architecture} illustrates the various augment flows that we classify them into: Low (L), Middle (M), High (H), L \(\times\) M, L \(\times\) H, M \(\times\) H, L \(\times\) M \(\times\) H, L \(+\) M, L \(+\) H, M \(+\) H, L \(+\) M \(+\) H. Here the symbol \(\times\) denotes a flow that undergoes multiple levels SC (Cascaded), while the symbol \(+\) represents the augmented features obtained from different levels considered as individual output flow (Pralleled). Our definitions of Low, Middle, and High levels can be found on the left of Figure \ref{fig: architecture}. For a discussion on this part of the motivation, please refer to the \textit{Supplementary Materials}.\par

In order to further reduce the sensitivity to style shift, we introduce the Channel-Wise Style Attention, which is a SE-like \cite{hu2018squeeze} module. First, calculating the mean \(\mu\) and variance \(\sigma^2\) of each channel within the feature map. These channel-wise styles are aggregated through a nonlinear operation, and finally, the feature response of each channel is obtained through a sigmoid activation:
\begin{equation}
\label{eqn: cwsa}
a=\mathrm{Sigmoid}(W_2(\mathrm{ReLU}(W_1(\mathrm{cat}(\mu_X,\sigma_X^2))))),
\end{equation}
\begin{equation}
\hat{X}=a\cdot X + X,
\end{equation}

Adaptive scaling those channels with domain-specific style variation, thereby maximizing the extraction of domain-independent valid information and mitigating the adverse impact of style shift.

\subsection{Asymmetric Augment Instance Contrast}
\label{Asymmetric Augment Instance Contrast}
\begin{figure}[t!]
\includegraphics[width=0.4\textwidth,height=0.15\textwidth]{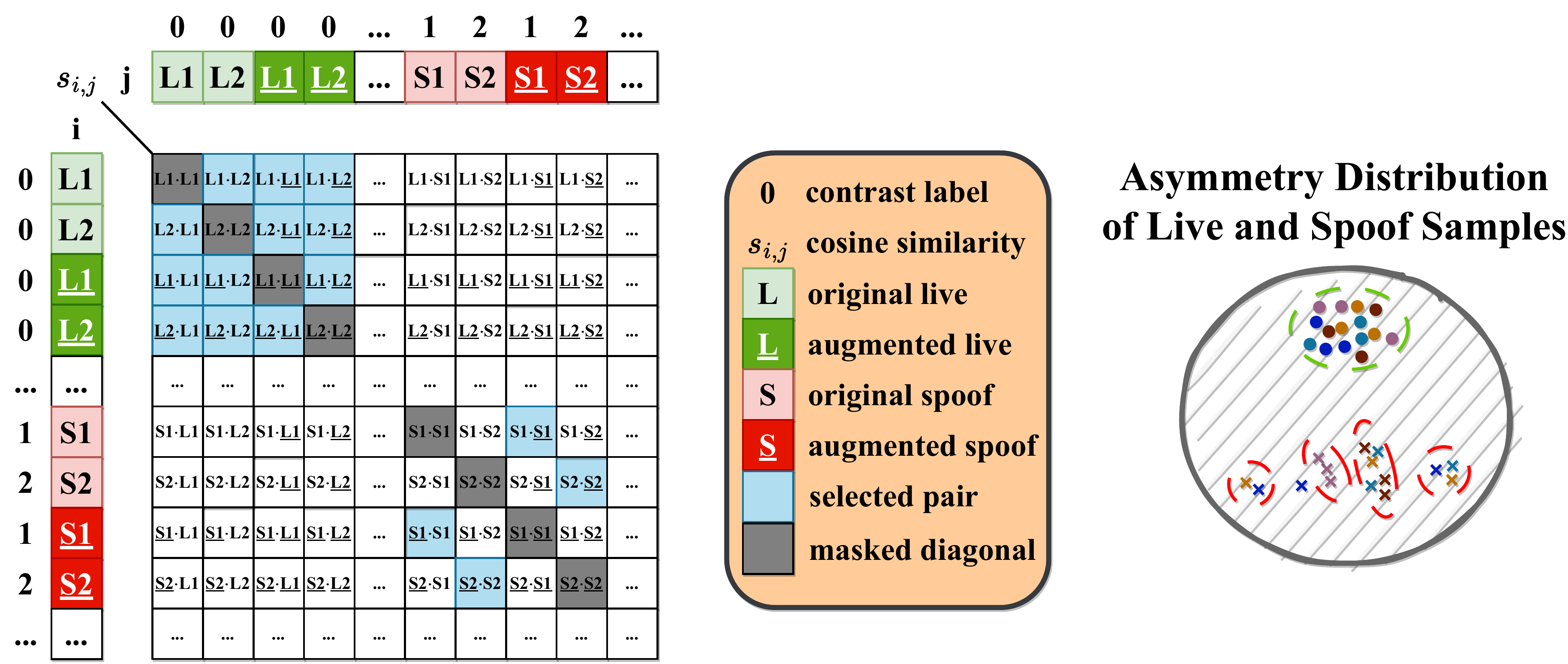}
\centering
\caption{AAIC results in a compact cluster of live samples, scatter pattern of spoofs.}
\label{fig: AAIC}
\vspace{0.5cm}
\end{figure}

When considering the FAS problem from the perspective of structural materials, it is observed that all live samples exhibit highly similar surface materials and surface reflectance properties. In contrast, spoofs exhibit greater diversity. Homogenize all spoof samples (Binary Contrast) or treating spoofs with dataset-gap-aware (Asymmetric Triplet) is obviously not optimal, but if each type of attack is labeled to perform a fine attack discrimination task, that will increase labor costs and is not beneficial for the generalization to unseen attacks. To address this issue, we aim to exclusively maximize the similarity between the original version spoofs and its Style Cross version spoofs within a batch, due to their liveness representations should be more consistent compared to other samples in the same batch (instance-pair-aware). Concurrently, we bring all original and augmented live instances closer together. This dual focus not only accommodates the asymmetry but also mitigates the model's sensitivity to the stylistic variations, as shown in Figure \ref{fig: AAIC}. We term our contrast strategy as asymmetric augmented instances contrast (AAIC). The AAIC is defined in Equation \ref{eqn: loss aaic}:
\begin{equation}
\label{eqn: loss aaic}
    L_{aaic}=-\frac{1}{B}\sum_i^{B} \sum_{\substack{i \neq j \\ y^i_c=y^j_c}}^B 
\frac{\exp(s_{i,j}/\tau)}{\sum_{\substack{i \\ i \neq k}}^{B} \exp(s_{i,k}/\tau)}%$
\end{equation}
where  $\tau$ is a hyper parameter, $y_c$ represents the contrast label. Elements on the diagonal are masked. For the FR task, we assign the same contrast labels to both the original and augmented samples from identical identities.

\subsection{Overall Objective Function}
For the FAS task, we use the asymmetric am-softmax loss (denoted as aams) in \cite{wang2022patchnet} as follows:

\begin{equation}
\label{eqn: loss cls}
L_{cls}=\frac{1}{B}\sum_i^B L_{aams}(C(\mathrm{Norm}(f_u^i)),y^i_{fas}),
\end{equation}
Then the total loss function of the FAS task is:
\begin{equation}
\label{eqn: loss u}
\small
L_{FAS}=L_{cls}+\lambda_{aaicU}L_{aaicU}+\lambda_{idamb}L_{idamb}+\lambda_{orthoU}L_{orthoU},
\end{equation}
For the FR task, we use the cross-entropy loss:
\begin{equation}
\label{eqn: loss id}
L_{id}= \frac{1}{B}\sum_i^B \mathrm{CrossEntropy}(D(\mathrm{Norm}(f_v^i)),y^i_{id}),
\end{equation}
the total loss function of the FR task is:
\begin{equation}
\label{eqn: loss v}
\small
L_{FR}=L_{id}+\lambda_{aaicV}L_{aaicV}+\lambda_{liamb}L_{liamb}+\lambda_{orthoV}L_{orthoV},
\end{equation}
% It is worth noting that in the orthogonal and contrast loss, the augmented instances are involved, while in the classification and ambiguity loss, we only deliver the original instances to the classifier or discriminator for the gradient backward. 
The optimization process is presented in Algorithm \ref{alg:1}.

\section{Experiment}
\subsection{Datasets and Protocol}
We use four public datasets: Oulu-NPU (denoted as O) \cite{boulkenafet2017oulu}, CASIA-FASD (denoted as C) \cite{zhang2012face}, Idiap Replay-Attack(denoted as I) \cite{chingovska2012effectiveness}, MSU-MFSD (denoted as M) \cite{wen2015face}, and follow the cross-datasets protocol same as previous DG-based methods \cite{jia2020single,chen2021generalizable,wang2022domain,wang2022patchnet,zhou2023instance,sun2023rethinking,liu2023towards} to evaluate the effectiveness of our method. The evaluation metrics are Half Total Error Rate (HTER) and Area Under the Curve (AUC).
% We also evaluate our approach under the intra-dataset protocols. Please refer to the supplementary materials for details. 

\subsection{Implementation Details}
We employ ResNet-18 \cite{he2016deep} as our backbone. Utilizing MTCNN \cite{zhang2016joint} for face detection, followed by cropping and resizing the facial area to a size of 256 \(\times\) 256. We apply random resized cropping and rotation for data augmentation. In training, each batch involves the random selection of four different IDs from each dataset and randomly choosing four live faces and four spoof faces from each ID. We use Adam optimizer \cite{kingma2014adam}, and the initial learning rate is set to 0.0005, which is decayed by 2 after every 50 epochs, the total training epoch is 200. We set the weight-decay as 5e-4, $\tau$=0.07, and set \(\lambda_{aaicU}\)=\(\lambda_{idamb}\)=\(\lambda_{orthoU}\)=\(\lambda_{aaicV}\)=\(\lambda_{liamb}\)=\(\lambda_{orthoV}\)=1, $U$, $C$, $V$, $D$ share the same hyper-parameter, optimizer, learning rate decay scheduler. Our method is implemented under the Pytorch framework.

% Suppose that we have three source datasets, the equivalent batch size is 96. 

\begin{algorithm}[t!]
\renewcommand{\algorithmicrequire}{\textbf{Input:}}
\renewcommand{\algorithmicensure}{\textbf{Output:}}
\caption{The  optimization strategy of \textbf{DLIF} network}
\label{alg:1}
\begin{algorithmic}[1]
\REQUIRE Source Data $S=\{x^j,y_{fas}^j,y_{id}^j\}_{n_s,n_{id}}^j$,Target Data
$T=\{x^j,y_{fas}^j\}_{n_t}^j$ , $U_\theta(\cdot)$,$C_\theta(\cdot)$,$V_\theta(\cdot)$,$D_\theta(\cdot)$
\ENSURE $U_{\theta^*}(\cdot)$,$C_{\theta^*}(\cdot)$,$V_{\theta^*}$,$D_{\theta^*}(\cdot)$
\WHILE {not end of iteration}
\STATE Sampling a mini-batch $B$ samples with $M$ identities: $X_s=\{x^i,y_{fas}^i,y_{id}^i\}_{B,M}^i$
\STATE For FAS task:$f_u^i=U_\theta(x^i),\hat{y}_{fas}^i=C_\theta(\mathrm{Norm}(f_u^i))$,
$f_{uAug}^i=U_\theta(LISC(x^i)),\hat{y}_{liamb}^i=D_\theta(\mathrm{Norm}(f_u^i))$
\STATE For FR task:$f_v^i=V_\theta(x^i),\hat{y}_{id}^i=D_\theta(\mathrm{Norm}(f_v^i))$,
$f_{vAug}^i=V_\theta(IISC(x^i)),\hat{y}_{idamb}^i=C_\theta(\mathrm{Norm}(f_v^i))$
\STATE Compute $L_{cls},L_{orthoU},L_{aaicU},L_{idamb}$, Update $\theta$ of $U$ and $C$ with Loss (\ref{eqn: loss u})
\STATE Compute $L_{id},L_{orthoV},L_{aaicV},L_{liamb}$, Update $\theta$ of $V$ and $D$ with Loss (\ref{eqn: loss v})
\STATE Evaluate $U_\theta(\cdot)$,$C_\theta(\cdot)$ on $T=\{x^j,y_{fas}^j\}_{n_t}^j$
\IF {performance better} 
\STATE Update $\theta^*$ of $U_{\theta^*}(\cdot)$,$C_{\theta^*}(\cdot)$,$V_{\theta^*}(\cdot)$,$D_{\theta^*}(\cdot)$ with $\theta$
\ENDIF
\ENDWHILE
\end{algorithmic}  
\textbf{Return} $U_{\theta^*}(\cdot)$,$C_{\theta^*}(\cdot)$,$V_{\theta^*}(\cdot)$,$D_{\theta^*}(\cdot)$
\end{algorithm}

\begin{table*}[t!]
\centering
\caption{Comparison to state-of-the-art FAS methods under the LOO setting. The bold numbers indicate the SOTA, while the underline indicates close to SOTA, second only to SOTA. HTER $\downarrow$ indicates smaller values are better, and AUC $\uparrow$ indicates larger values are better.}
\vspace{0.5cm}
\label{table: OCMI}
\resizebox{0.9\textwidth}{!}{
\begin{tabular}{c|c|c|c|c|c|c|c|c}
\cmidrule(r){1-9}
\multirow{2}{*}{Method} & \multicolumn{2}{c|}{O\&C\&I to M} & \multicolumn{2}{c|}{O\&M\&I to C} & \multicolumn{2}{c|}{O\&C\&M to I} & \multicolumn{2}{c}{I\&C\&M to O} \\
\cmidrule(r){2-3} \cmidrule(r){4-5} \cmidrule(r){6-7} \cmidrule(r){8-9}
& HTER $\downarrow$ (\%) & AUC $\uparrow$ (\%) & HTER $\downarrow$ (\%) & AUC $\uparrow$ (\%) & HTER $\downarrow$ (\%) & AUC $\uparrow$ (\%) & HTER $\downarrow$ (\%) & AUC $\uparrow$ (\%) \\
\cmidrule(r){1-9}
MMD-AAE \cite{li2018domain} & 27.08 & 83.19 & 44.59 & 58.29  & 31.58 & 75.18 & 40.98 & 63.08 \\
MADDG \cite{shao2019multi} & 17.69 & 88.06 & 24.50 & 84.51 & 22.19 & 84.99 & 27.98 & 80.02 \\
NAS-FAS \cite{yu2020fas} & 19.53 & 88.63   & 16.54 & 90.18  & 14.51 & 93.84  & 13.80 & 93.43 \\
RFM \cite{shao2020regularized} & 13.89 & 93.98   & 20.27 & 88.16  & 17.30 & 90.48  & 16.45 & 91.16 \\
SSDG \cite{jia2020single} & 7.38  & 97.17  & 10.44  & 95.94 & 11.71  & 96.59 & 15.61  & 91.54 \\
D\(^2\)AM \cite{chen2021generalizable} & 12.70  & 95.66  & 20.98  & 85.58 & 15.43  & 91.22 & 15.27  & 90.87 \\
DRDG \cite{liu2021dual} & 12.43  & 95.81  & 19.05  & 88.79 & 15.56  & 91.79 & 15.63  & 91.75 \\
ANRL \cite{liu2021adaptive} & 10.83  & 96.75  & 17.83  & 89.26 & 16.03  & 91.04 & 15.67  & 91.90 \\
SSAN \cite{wang2022domain} & 6.67  & \textbf{98.75}  & 10.00  & 96.67 & 8.88  & 96.79 & 13.72  & 93.63 \\
PatchNet \cite{wang2022patchnet} & 7.10  & 98.46  & 11.33  & 94.58 & 13.40  & 95.67 & 11.82  & 95.07 \\
IADG \cite{zhou2023instance} & 5.41  & 98.19  & 8.70  & 96.44 & 10.62  & 94.50 & \textbf{8.86}  & \textbf{97.14} \\
SA-FAS \cite{sun2023rethinking} & 5.95  & 96.55  & 8.78  & 95.37 & 6.58  & 97.54 & 10.00  & 96.23 \\
UDG-FAS \cite{liu2023towards} & 5.95  & 98.47  & 9.82  & 96.76 & \underline{5.86}  & \textbf{98.62} & 10.97  & 95.36 \\
\cmidrule(r){1-9}
\textbf{DLIF (Ours)} & \textbf{3.75}  & 98.33  & \textbf{6.67}  & \textbf{97.27} & \textbf{5.82}  & \underline{98.13} & \underline{8.89} & \underline{96.36} \\
\cmidrule(r){1-9}
\end{tabular}}
\end{table*}

\begin{table}[h!]
\caption{Comparison under limited source domains setting}
\vspace{0.5cm}
\label{table: MI}
\resizebox{0.45\textwidth}{!}{
\begin{tabular}{cccccccc}
\hline
\multirow{2}{*}{Method} & \multicolumn{2}{c}{M\&I to C} & \multicolumn{2}{c}{M\&I to O} \\
\cmidrule(r){2-3} \cmidrule(r){4-5}
& HTER $\downarrow$ (\%) & AUC $\uparrow$ (\%) & HTER $\downarrow$ (\%) & AUC $\uparrow$ (\%)  \\
\hline
MADDG \cite{shao2019multi} & 41.02 & 64.33 & 39.35 & 65.10  \\
SSDG-M \cite{jia2020single} & 31.89  & 71.29  & 36.01  & 66.88 \\
D\(^2\)AM \cite{chen2021generalizable} & 32.65  & 72.04  & 27.70 & 75.36 \\
DRDG \cite{liu2021dual} & 31.28  & 71.50  & 33.35  & 69.14 \\
ANRL \cite{liu2021adaptive} & 31.06  & 72.12  & 30.73  & 74.10 \\
SSAN-M \cite{wang2022domain} & 30.00  & 76.20  & 29.44  & 76.62 \\
IADG \cite{zhou2023instance} & 24.07  & 85.13  & 18.47  & \textbf{90.49} \\
\hline
\textbf{DLIF (Ours)} & \textbf{11.11}  & \textbf{92.77}  & \textbf{16.97}  & \underline{89.64} \\
\hline
\end{tabular}
}
\end{table}

\subsection{Comparison with state-of-art methods}
In accordance with the commonly applied protocols used in DG-FAS methods, we perform the Leave-One-Out (LOO) and limited source domains evaluation protocols.
\subsubsection{Leave-One-Out (LOO).}
In the LOO setting, we employ O, C, M, and I datasets, randomly selecting three of them as source domains for training, while the remaining one is held as the unseen target domain for testing. As shown in Table \ref{table: OCMI}, our method demonstrates superior performance compared to most of the other methods that use datasets as domain concepts. These results demonstrate the generalization ability of our method since we employ identity partition, which is more fine-grained and shrinks the scope of the domain. It is conducive to learning domain-invariant representations, as using the dataset as the domain partition will bring more significant intra-domain variations. Compared to SSAN, we propose a task-oriented style cross that ensures the rationality of style augmentation, and also attempt various style augmentation flows. In addition, our proposed AAIC loss is more effective than previous binary contrastive loss and triplet loss.

\subsubsection{Limited source domains.} As shown in Table \ref{table: MI}, we evaluate our method under the limited source domains. Following prior research, we select M and I as source domains, while C and O, are respectively utilized as the unseen target domain. Our method achieves a significant improvement compared to previous state-of-the-art methods. It is proved that eliminating the effect of identity shift and employing a finer domain partition is extremely effective for the unseen target domain in the case of limited source data and source identities.

\subsection{Ablation and Discussion}
In this subsection, we conduct ablation studies on individual contribution of components. Additionally, we compare various contrast losses and style augmentation strategies. All ablation and comparison studies are performed under the O\&C\&I to M setting.

\begin{table}[t!]
\caption{Ablation of each component on O\&C\&I to M}
\vspace{0.5cm}
\label{table: Ablation}
\resizebox{0.47\textwidth}{!}{
\begin{tabular}{cccccccc}
\hline
Baseline & V & CWSA & SC (M+H) & HTER $\downarrow$ (\%) & AUC $\uparrow$ (\%)  \\
\hline
\checkmark & - & - & - & 12.08  & 94.21 \\
\checkmark & - & - & \checkmark (LISC) & 9.17  & 97.31 \\
\checkmark & - & \checkmark & - & 10.00  & 96.14 \\
\checkmark & \checkmark & - & - & 7.91  & 95.79 \\
\checkmark & \checkmark & - & \checkmark (LI-IISC) & 8.66  & 96.92 \\
\checkmark & \checkmark & \checkmark & - & 7.92  & 97.08 \\
\checkmark & - & \checkmark & \checkmark (LISC)  & 6.70  & 97.83 \\
\checkmark & \checkmark & \checkmark & \checkmark (LI-IISC) & \textbf{3.75}  & \textbf{98.33} \\
\hline
\end{tabular}
}
\end{table}

\subsubsection{Contribution of each component.} 
Table \ref{table: Ablation} shows the ablation studies. For the baseline configuration, we employ ResNet-18 as the backbone. Only aam-softmax loss and binary contrast loss are utilized, without any additional components. Given that there are three components in total, a total of seven combinations are formed. We conduct experiments on all combinations, and M+H flow in SC is consistently used. In the case of introducing SC, the contrast loss for the FAS task is replaced with AAIC. Furthermore, the introduction of \(V\) involves employing the orthogonal loss for both \(U\) and \(V\). The results show that incorporating each component individually leads to performance improvements compared to the baseline. Moreover, when including two components simultaneously, further enhancements are observed. Notably, our model achieves state-of-the-art when all three components are integrated concurrently, thus confirming the effectiveness of each component.

\subsubsection{Comparisons of different contrast strategies.}
The performance of various contrast losses is presented on the upper side of Table \ref{table: contrast and augmentation}. In comparison to Binary and Triplet, AAIC achieves superior generalization ability, because the Binary disregards the inherent differences in density and diversity between live and spoof samples. Although Triplet notices this distinction which is better than Binary, the domain-gap-aware is not fine-grained enough. In contrast, AAIC focuses on asymmetry, ie., considering that live instances exhibit homogeneity while spoof instances display heterogeneity, besides, adopting instance-pair-aware is more refined than domain-gap-aware for spoofs.

\begin{table*}[t!]
\centering
\caption{Scalability 1) : Effectiveness of well-trained FR networks. * means that the encoder \(V\) with pre-trained weights. \textbf{Bold} represents the best result. HTER $\downarrow$ indicates smaller values are better, and AUC $\uparrow$ indicates larger values are better. \textcolor{red}{Red} font represents ascent and \textcolor{blue}{Blue} font indicates descent.}
\vspace{0.5cm}
\label{table: OCMI+}
\resizebox{0.75\textwidth}{!}{
\begin{tabular}{ccccccccc}
\hline
\multirow{2}{*}{Backbone} & \multicolumn{2}{c}{O\&C\&I to M} & \multicolumn{2}{c}{O\&M\&I to C} & \multicolumn{2}{c}{O\&C\&M to I} & \multicolumn{2}{c}{I\&C\&M to O} \\
\cmidrule(r){2-3} \cmidrule(r){4-5} \cmidrule(r){6-7} \cmidrule(r){8-9}
& HTER 
$\downarrow$ (\%)& AUC $\uparrow$ (\%)& HTER $\downarrow$ (\%)& AUC $\uparrow$ (\%)& HTER $\downarrow$ (\%)& AUC $\uparrow$ (\%)& HTER $\downarrow$ (\%)& AUC $\uparrow$ (\%)\\
\hline
ResNet18 & 3.75  & 98.33  & 6.67  & 97.27& 5.82  & 98.13& 8.89  & 96.36 \\
\hline
FaceNet & 5.47  & 97.22  & 8.70  & 95.14 & 10.17  & 94.51 & 9.49  & 94.86 \\
FaceNet* & \textcolor{blue}{3.13 (-2.34)}  & \textcolor{red}{98.45 (+1.23)} & 8.88  & \textcolor{red}{96.40 (+1.26)} & \textcolor{blue}{5.52 (-4.65)} & \textcolor{red}{97.50 (2.99)} & \textcolor{blue}{9.03 (-0.46)}  & \textcolor{red}{96.17 (+1.31)} \\
\hline
CosFace & 4.91  & 98.16  & 6.66 & 96.58 & 9.64  & 95.55 & 9.27 & 96.60 \\
CosFace* & \textcolor{blue}{4.30 (-0.61)}  & \textcolor{red}{98.68 (+0.52)} & \textcolor{blue}{5.26 (-1.4)} & \textcolor{red}{97.04 (+0.46)} & \textcolor{blue}{7.43 (-2.21)} & 95.39 & \textcolor{blue}{8.64 (-0.63)} & \textcolor{red}{97.11 (+0.51)} \\
\hline
\end{tabular}}
\end{table*}

\begin{table*}[t!]
\centering
\caption{Scalability 2) : Effectiveness of identity diversity. Gray fill indicates the baseline. HTER $\downarrow$ indicates smaller values are better, and AUC $\uparrow$ indicates larger values are better. \textcolor{red}{Red} font represents ascent and \textcolor{blue}{Blue} font indicates descent.}
\vspace{0.5cm}
\label{table: CS-OCMI}
\resizebox{0.75\textwidth}{!}{
\begin{tabular}{ccccccccc}
\hline
\multirow{2}{*}{Backbone} & \multicolumn{2}{c}{O\&C\&I to M} & \multicolumn{2}{c}{CS\&C\&I to M} & \multicolumn{2}{c}{O\&CS\&I to M} & \multicolumn{2}{c}{O\&C\&CS to M} \\
\cmidrule(r){2-3} \cmidrule(r){4-5} \cmidrule(r){6-7} \cmidrule(r){8-9}
& HTER $\downarrow$ (\%)& AUC $\uparrow$ (\%)& HTER $\downarrow$ (\%)& AUC $\uparrow$ (\%)& HTER $\downarrow$ (\%)& AUC $\uparrow$ (\%)& HTER $\downarrow$ (\%)& AUC $\uparrow$ (\%)\\
\hline
\textbf{DLIF} (FaceNet*) & \cellcolor{gray}3.13  & \cellcolor{gray} 98.45  & \textcolor{blue}{2.42 (-0.71)}  & \textcolor{red}{99.44 (+0.99)} & \textcolor{blue}{1.79 (-1.34)}  & \textcolor{red}{99.67 (+1.22)}  & \textcolor{blue}{1.84 (-1.29)} & \textcolor{red}{99.69 (+1.24)}  \\
\textbf{DLIF} (CosFace*) & \cellcolor{gray} 4.30  & \cellcolor{gray} 98.68  & \textcolor{blue}{1.92 (-2.38)} & \textcolor{red}{99.68 (+1.00)} & \textcolor{blue}{2.03 (-2.27)} & \textcolor{red}{99.52 (+0.84)} & \textcolor{blue}{1.64 (-2.66)} & \textcolor{red}{99.81 (+1.13)}  \\
\hline
\end{tabular}}
\end{table*}

\subsubsection{Comparisons of different style augmentation strategies.}
Table \ref{table: SC level} shows the impact of implementing SC at various levels (L, M, H) and employing different augmentation flows (\(\times, +\)). Several key findings can be drawn from the results: 1) Solely Performing SC at a low level could lead to a slight performance decrease. 2) Irrespective of whether it is \(\times\)or \(+\), the multi-level SC is more effective than the single-level, and the combination M+H yields the best results. Additionally, the lower side of Table \ref{table: contrast and augmentation} reveals that LISC outperforms SSA, indicating that LISC is more suitable for the FAS task than SSA. Moreover, implementing the IISC in \(V\) is also indispensable. The unbalanced expansion of the $\mathcal{U}$ space alone may cause the $\mathcal{V}$ space to be squeezed, which may lead to the leakage of certain identity information.

% \begin{figure}[t!]
%     \centering
% \includegraphics[width=0.37\textwidth,height=0.22\textwidth]{Scalability.pdf}
% \caption{Scalable framework.}
% \label{fig: Framework}
% \end{figure}

\begin{table}[t!]
\centering
\caption{Comparison of contrast and augment on O\&C\&I to M}
\vspace{0.5cm}
\label{table: contrast and augmentation}
\resizebox{0.43\textwidth}{!}{
\begin{tabular}{ccccccccc}
\hline
{Method} & {Option} & {HTER $\downarrow$ (\%)} & {AUC $\uparrow$ (\%)}\\
\hline
\multirow{3}{*}{Contrast Strategy} & Binary & 9.70 & 97.31 \\ 
                                                                \cmidrule(r){2-4}
                                                                & Triplet & 5.42 & 96.85\\
                                                                \cmidrule(r){2-4}
                                                                & \textbf{Ours(AAIC)} & \textbf{3.75} & \textbf{98.33} \\
\hline
\multirow{3}{*}{Augment Strategy} & SSA & 10.00 & 95.54\\
                                                                  \cmidrule(r){2-4}
                                                                  & Only LISC & 7.41 & 95.60\\
                                                                  \cmidrule(r){2-4}
                                                                  & LISC \& IISC  & \textbf{3.75} & \textbf{98.33} \\
\hline
\end{tabular}
}
\end{table}

\subsubsection{Scalability Study.}
% In this subsection, we explore our framework's scalability. For more detailed information on this section, please refer to the supplementary materials. datasets such as LFW, IJB-C, and CASIA-WebFace provide abundant identity information for training a network with a strong ability to extract identity features.  Importantly, our framework does not need to label these identities.
Benefiting from the success of face recognition, numerous well-trained open-source FR models are available, then \(V\) no longer needs to be trained from scratch. In this regard, we delve into how our model can benefit from a well-trained FR network, given that we requires identity features in essence, not identity labels. We outline the scalability as follows: \textbf{Scalability 1):} Leveraging well-trained FR networks to provide disentanglement guidance for FAS networks. We replace the ResNet18 with FaceNet and CosFace, respectively. In Table \ref{table: OCMI+}, we employ them as the backbone of encoders \(U\) and \(V\). The results demonstrates that if well-trained weights are loaded, most protocols have improved under the LOO setting. This suggests that our framework is compatible with most FR models, highlighting the advantages of utilizing well-trained FR models to extract identity representations. In addition, concerning training efficiency, utilizing ResNet18 as the backbone for two encoders, requiring only 2-3 hours to complete 200 epochs on two 1080Ti GPUs. Furthermore, freezing a well-trained encoder \(V\) during training can significantly improve training efficiency. \textbf{Scalability 2):} Increasing the diversity of identities in the source domain will significantly improve the performance. We conduct experiments to demonstrate that increasing the identity diversity in training phase will improve the generalization capability. Given the scarcity of identities in OCIM, where O has 40 IDs, C has 50 IDs, I has 35 IDs, and M has 55 IDs. However, CelebA Spoof (denoted as CS) is an open-source FAS dataset with identity diversity, consisting of 27260 IDs. The IDs in CS surpass the total number of OCIM identities by more than 150 times. The specific experimental setup is as follows: we sequentially replace the three source datasets in the O\&C\&I to M protocol with CS, designating M as the target domain dataset (excluding M because it contains the most IDs in OCIM). The results presented in Table \ref{table: CS-OCMI} highlight that increasing identity diversity significantly improves model performance.

 % In this way, we can acquire a large amount of unlabeled data from the internet or existing face recognition datasets, facilitating representation disentanglement.
% An asterisk (*) denotes the loading pretrained weights for face recognition to encoder \(V\), and frozen the weight of encoder \(V\) during training. In this case, an identity discriminator is not needed. Absence of an asterisk indicates that encoder \(V\) is trained from scratch. 

% due to our framework essentially requires auxiliary supervised identity features rather than identity labels

\begin{figure}[t!]
\includegraphics[width=0.48\textwidth,height=0.17\textwidth]{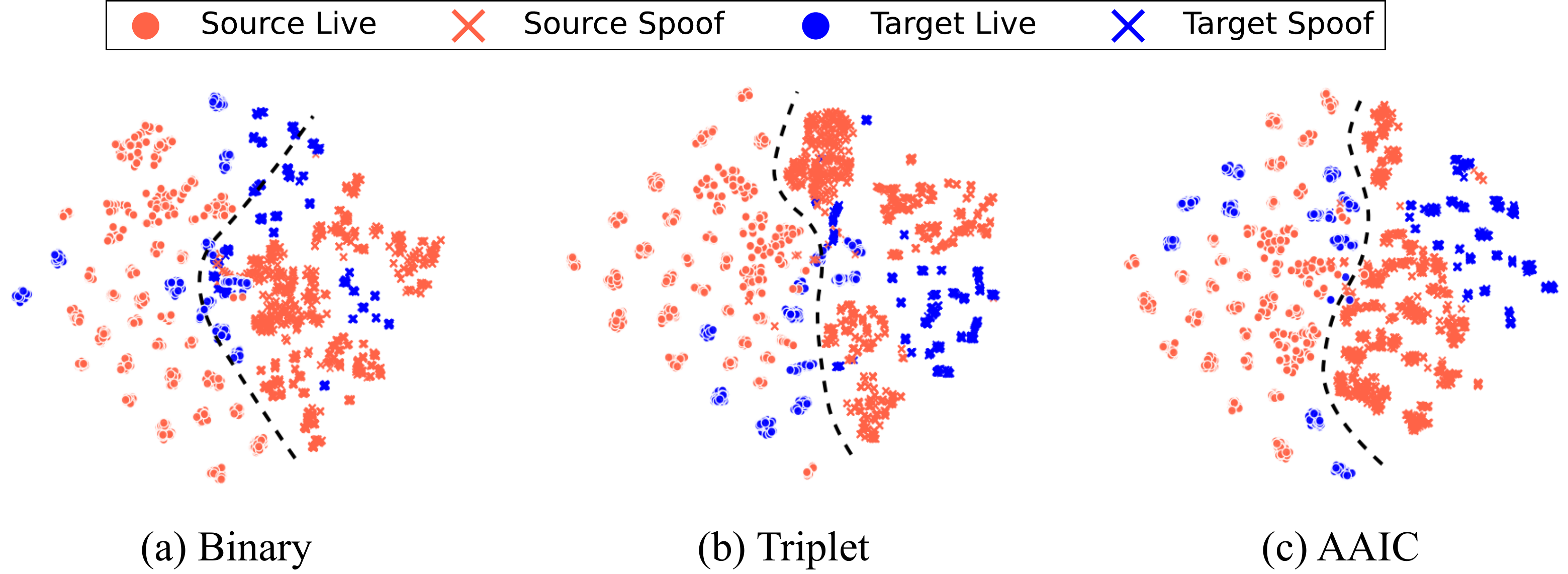}
\centering
\caption{Feature distribution of different contrast strategies via t-SNE visualization.}
\label{fig: contrast strategy}
\vspace{0.7cm}
\end{figure}

\begin{table*}[t!]
\centering
\caption{Ablation of different SC flows on O\&C\&I to M. The bold indicates the best and the underlined means the worst}
\vspace{0.5cm}
\label{table: SC level}
\resizebox{0.9\textwidth}{!}{
\begin{tabular}{cccccccccccc}
\hline
Level & L & M & H & L \(\times\) M & L \(\times\) H & M \(\times\) H & L \(\times\) M \(\times\) H & L + M & L + H & M + H & L + M + H\\
\cmidrule(r){1-1} \cmidrule(r){2-2} \cmidrule(r){3-3} \cmidrule(r){4-4} \cmidrule(r){5-5} \cmidrule(r){6-6} \cmidrule(r){7-7} \cmidrule(r){8-8} \cmidrule(r){9-9} \cmidrule(r){10-10} \cmidrule(r){11-11} \cmidrule(r){12-12}
HTER $\downarrow$ (\%)& \underline{12.92} & 10.00 & 7.92 & 5.00 & 6.80 & \textbf{5.00} & 7.08 & 10.21 & 7.50 & \textbf{3.75} & 7.50 \\
AUC $\uparrow$ (\%)& \underline{93.97} & 94.03 & 96.13 & 97.92 & 97.82 & \textbf{98.00} & 98.19 & 97.02 & 97.51 & \textbf{98.33} & 97.12 \\
\hline
\end{tabular}
}
\end{table*}

\begin{figure*}[!]
\includegraphics[width=0.83\textwidth,height=0.22\textwidth]{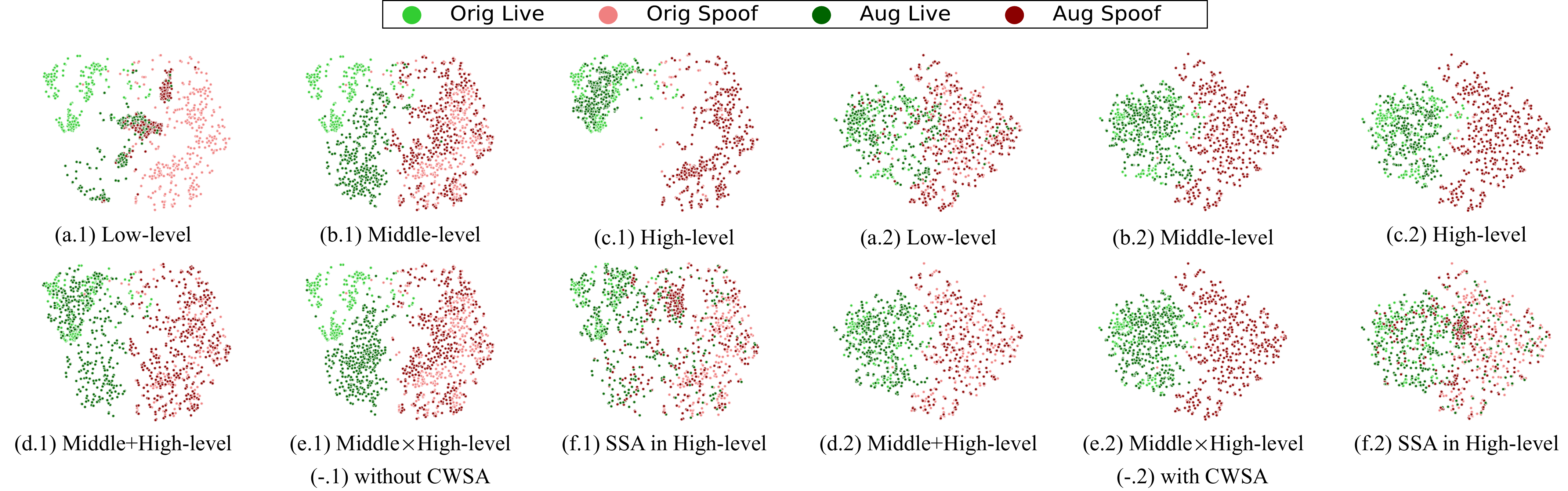}
\centering
\caption{(a.-) (b.-), (c.-), (d.-), (e.-), and (f.-) correspond to the feature distribution of L, M, H, M+H, M \(\times\)H, SSA these six style augmentation methods respectively. The (-.1) and (-.2) indicate whether \(U\) is equipped with the CWSA.}
\label{fig: SC aug flow}
\end{figure*}

\section{Visualization and Analysis}
\subsection{Feature distributions of different contrast losses.}
The distribution of features optimized by different contrast loss is visualized by t-SNE \cite{van2008visualizing}, as shown in Figure \ref{fig: contrast strategy}. In (a), the source spoofs are compact, while the target spoofs appear relatively scattered. Regarding target samples, the distance between live and spoof samples is closer. In (b), source spoofs cluster into three cliques, but the target spoofs do not belong to any of them, and the distance between the live and spoof samples of the target is not far enough. For (c), the spoof distribution is loose, and the distance between the live and spoof samples of the target is farther than the (a) and (b). The reason is that AAIC is equivalent to adopting a more refined clustering for spoofs, where each spoof instance pair has a unique contrast label in a batch, resulting in a more dispersed distribution of spoofs, while the instance-pair-aware can also weaken the style sensitivity.

\begin{figure}[t!]
\centering
\includegraphics[width=0.48\textwidth,height=0.26\textwidth]{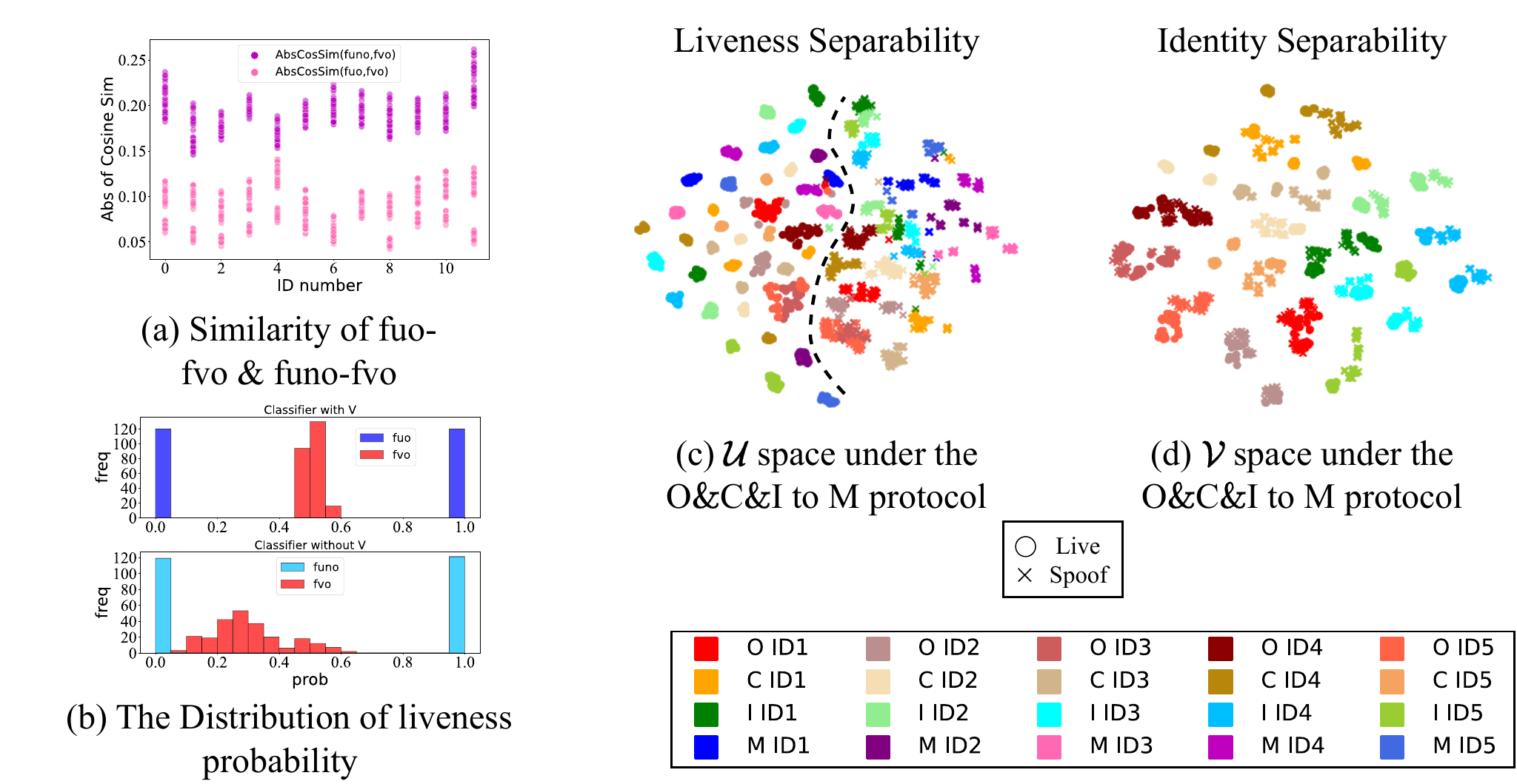}

\caption{The orthogonality of space $\mathcal{U}$ \& $\mathcal{V}$.}
\label{fig: Orthogonality}
\vspace{0.7cm}
\end{figure}

\subsection{Orthogonality derives from orthogonal constrain.} To prove that introducing \(V\), orthogonal, ambiguous loss can help \(U\) learn identity-invariant representations, we use \(U_o\), \(C_o\), \(V_o\) which are optimized simultaneously with orthogonal and ambiguous loss, another \(U_{no}\), \(C_{no}\) having the same setting as \(U_o\), \(C_o\) except for the orthogonal and ambiguous constraint. As shown in Figure \ref{fig: Orthogonality} (a), we randomly select 12 IDs from source data and compute the absolute value of cosine similarity of $f_{vo}$ and $f_{uo}$ or $f_{uno}$. The points of each column represent the similarity between the $f_{vo}$ of the corresponding abscissa's ID and the $f_{uo}$ (pink points) or $f_{uno}$ (purple points) of all IDs. Due to orthogonal loss and FR task-relevant loss, \(V_o\) can get identity attributes and filter out liveness attributes as much as possible, thus the pink points are smaller around 0.075, but purple points are larger around 0.2. The upper subgraph of Figure \ref{fig: Orthogonality} (b) shows the \(C_o(\mathrm{Norm}(f_{vo}))\) (liveness probability) concentrated around 0.5, and \(C_o(\mathrm{Norm}(f_{uo}))\) locates in 0 or 1, however,  the lower subgraph of Figure \ref{fig: Orthogonality} (b) shows the \(C_{no}(\mathrm{Norm}(f_{vo}))\) is more scattered, and \(C_{no}(\mathrm{Norm}(f_{uno}))\) locates in 0 or 1. This phenomenon proves that the classifier without introducing orthogonal and ambiguous loss does not exhibit liveness ambiguity to \(f_{vo}\). As shown in Figure \ref{fig: Orthogonality} (c), (d), colorful points represent different identities from various datasets, we can observe that $\mathcal{U}$ space has liveness separability and $\mathcal{V}$ space has identity separability when introducing the orthogonality and ambiguity. 

\subsection{The effectiveness of different SC flows and CWSA.}
We visualize the all augmented features obtained from various style augmentation flows alongside the original features in the same coordinate system by t-SNE. As shown in Figure \ref{fig: SC aug flow} (-.1), where are not equipped with CWSA, the following three results can be observed: 1) multiple levels style cross yields a more style diverse feature, as in (d.1) (e.1) versus (a.1) (b.1) (c.1). 2) In (a.1) we find that the augmented features approach or even cross the classification hyperplane, such that L-level SC degrades the performance. Thus for \(+\) and \(\times\) Aug Flow, augmentation without L is preferable indicating that style cross is recommended at middle and high levels. 3) As shown in (f.1), we can observe the liveness variation, which can explain why LISC is more suitable for the FAS task than SSA. Moreover, in Figure \ref{fig: SC aug flow} (-.2), after equipping with CWSA, similar feature distributions were ultimately obtained for different style augmentation flows. This indicates that CWSA has the robustness to diverse style shifts.

\begin{figure}[t!]
\includegraphics[width=0.46\textwidth,height=0.13\textwidth]{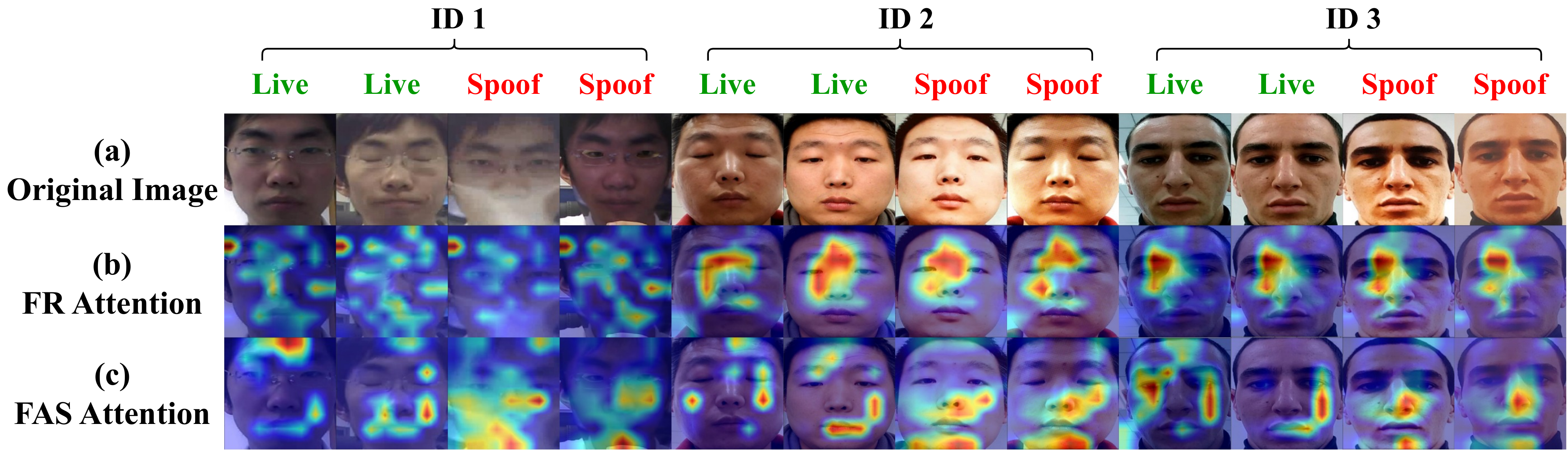}
\centering
\caption{Grad-CAM visualizations of activation areas.}
\label{fig: GradCAM}
\vspace{0.7cm}
\end{figure}

\subsection{Attention visualization of GradCAM.}
We use Grad-CAM \cite{selvaraju2017grad} to visualize the activation map of the last layer of  \(U\) and \(V\). In Figure \ref{fig: GradCAM}, for faces with the same identity, the regions of \textbf{FR} attention are similar, while the difference in \textbf{FAS} attention regions is reflected in liveness, and it can be observed that \textbf{FAS} attention exhibits overlaps in certain extent for live samples and highly overlapping for spoofs with similar attack patterns.

\section{Conclusion}
In this work, we propose a novel perspective to learn identity-invariant spoof representations, by simultaneously training two networks for the FAS task and the FR task and constraining them through orthogonal loss to disentangle the liveness and identity. We also utilize the task-oriented style augmentation, as well as the CWSA to weaken the style sensitivity and design the AAIC. Through extensive experiments, we demonstrate the effectiveness of our method which achieves SOTA performance on prevalent benchmarks. Especially in the case of limited source data, the advantage is obvious. Furthermore, our method has strong scalability.

\section{Acknowledgments}
% This research was supported by the advanced computing resources provided by the Supercomputing Center of the University of Science and Technology of China (USTC).
The numerical calculations in this paper have been done on the supercomputing system in the Supercomputing Center of University of Science and Technology of China.

%%% Use this command to include your bibliography file.
\bibliography{mybibfile}

\end{document}